%% file: example.tex
\pdfoutput=1 
\documentclass{article}
\usepackage[preprint]{corl_2025} 
\usepackage{graphicx}
\usepackage{bm} 
\usepackage{amsmath} 
\usepackage{amssymb}  
\usepackage{comment} 
\usepackage{cite}
\usepackage{multirow}
\usepackage{diagbox}

\title{ Imitation Learning Based on Disentangled Representation Learning of Behavioral Characteristics  }

\author{
  Ryoga Oishi$^1$, Sho Sakaino$^2$, Toshiaki Tsuji$^1$\\
  $^1$Saitama University \quad $^2$University of Tsukuba\\
  \texttt{r.oishi@ms.saitama-u.ac.jp} \\
}

\begin{document}
\maketitle

\begin{abstract}
In the field of robot learning, coordinating robot actions through language instructions is becoming 
increasingly feasible. However, adapting actions to human instructions remains challenging, as such 
instructions are often qualitative and require exploring behaviors that satisfy varying conditions.
This paper proposes a motion generation model that adapts robot actions in response to 
modifier directives—human instructions imposing behavioral conditions—during task execution. 
The proposed method learns a mapping from modifier directives to actions by segmenting 
demonstrations into short sequences, assigning weakly supervised labels corresponding to specific 
modifier types. We evaluated our method in wiping and pick-and-place tasks. Results show that it 
can adjust motions online in response to modifier directives, unlike conventional batch-based 
methods that cannot adapt during execution.
\end{abstract}


\begin{figure}[htbp]
    \centering
    \includegraphics[width=11.0cm]{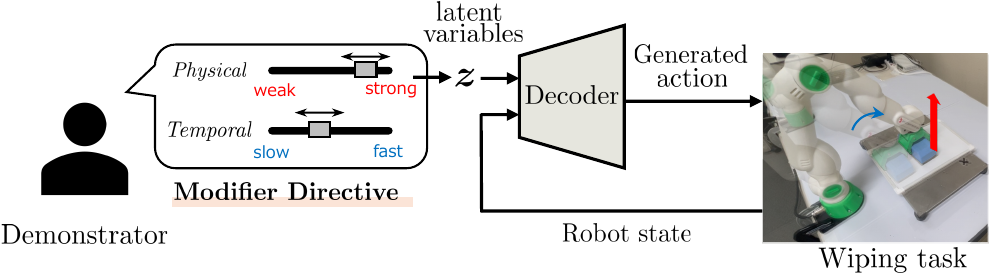}
    \caption{
    Overview of the proposed method. It generates the next motion trajectory based on a 
    human-given \textbf{modifier directive} and the current robot state. 
    In the example of wiping a whiteboard, a human gives \textbf{modifier directive} regarding 
    the strength and speed of wiping, and the robot generates the corresponding motion.}
    \label{fig:overview}
\end{figure}
\input{./src/01_intro.tex}

\input{./src/02_rerated_work.tex}
\input{./src/03_proposed.tex}
\input{./src/04_experiment.tex}
\input{./src/05_conclusion.tex}


\clearpage
\acknowledgments{
This paper is based on results obtained from a project,
JPNP14004, commissioned by the New Energy and Industrial Technology Development
Organization (NEDO). 
}


\bibliographystyle{unsrt}
\bibliography{example}  
\input{./src/06_appendix.tex}

\end{document}

%% file: src/01_intro.tex
\section{Introduction}
Imitation Learning (IL) \citep{osaIL} is a method that learns robot behavior based on human 
demonstration data and is positioned as one of the major approaches in robot learning, 
alongside reinforcement learning. IL is simple because it can learn a motion generation 
policy from demonstration data. It has been applied to various robot tasks such 
as folding tasks \citep{ogataral20217}, assembly tasks \citep{assembly_harada},
bottle-opening tasks \citep{takeuchiJIA}, and handwriting tasks 
\citep{kutsuzawaaccess}. However, learning an environment-adapted motion generation policy from 
limited demonstrations is challenging, and some studies have pointed out the limitations of 
learning motions solely from demonstrations \citep{Dorsa2022}. Therefore, many methods 
based on human intervention have been proposed as solutions to this issue \citep{ross2011, mandlekar2020}, 
with research advancing toward developing robots capable of executing complex tasks based on human 
instructions, linking human instructions to actions \citep{language-conditioned, Cliport}.

Many studies that aim to associate human instructions with robot motion generation focus on the 
combination of large language models and existing motion generation models \citep{stepputtis2020, bcz, 
kobayashi2025_bi-lat, saycan, rt2}. While these foundation models are highly effective for high-level 
task selection and coarse-grained skill switching, they still face significant challenges in performing 
online, fine-grained modulation of continuous motion parameters, such as speed, force, and smoothness. 
This limitation has also been highlighted in recent studies auditing robotic foundation 
models \citep{karnik2025embodied}. On the other hand, human instructions used for motion generation are often 
qualitative, making it difficult to establish a one-to-one mapping between the intended instruction and 
the resulting motion. For example, in the task of wiping a desk, instructions like ``wipe strongly'' 
or ``wipe weakly'' serve as modifiers that qualitatively adjust the force applied during motion execution.
In this paper, such instructions will be referred to as \textbf{modifier directives}. 
It is particularly challenging not only to learn the basic capability of generating 
a wiping motion, but also to adjust the force applied during the motion according to such directives
\citep{kawaharaduka2021}.


One method to directly link modifiers and learned representations is Disentanglement 
Representation Learning (DRL) \citep{bengio2014}. DRL aims to separate the factors contained in the 
learning data and encode them as independent, distinct latent variables in the representation 
space \citep{drl2018, wang2024drl}. Within DRL, methods have been proposed that impose constraints on the 
learning of the latent space to separate the desired representations \citep{2017betavae, 2019gatedvae}, 
and approaches for applying these methods to robot learning have also been proposed \citep{Peter2024disentangled}. 
An application to imitation learning involves the use of weakly supervised labels to impose constraints 
on the latent space representation \citep{hristov2021learning, SONG2023120625}. 
By assigning modifier directives as weakly supervised labels, it becomes possible to learn the 
correspondence between modifier directives and latent representations. By controlling these latent 
variables, robot motion trajectories corresponding to the modifier directives can be generated as 
outputs of the policy.

Many methods based on DRL for learning the correspondence between robot motions and modifier directives involve  
architectures that learn motions in a batch, and the application to online settings has not been explored. 
In real-world tasks, it is often expected that motion needs to be adjusted based on modifier directives during 
task execution, making it crucial to have a motion generation model capable of responding to such directives 
online. We define online motion generation as a process that dynamically adjusts a robot's motion in response 
to human-issued modifier directives during task execution. Achieving this requires a learning architecture 
that can learn the latent space corresponding to modifier directives and generate motions online based on 
that latent space.

This study proposes a motion generation model that allows robots to change their behavior online 
based on modifier directives. The study addresses the following two challenges: 
\textbf{1. Acquiring latent representations that link modifier directives and motions.}
 Generally, human motions targeted for imitation are high-dimensional, making direct control difficult. 
Therefore, it is desirable to manipulate motions through the latent variables of an autoencoder. 
However, these latent variables tend to lack interpretability, making it difficult to 
interpret and control them directly. 
To address this, we use DRL with weak supervision to learn a latent space corresponding to 
modifier directives, enabling motion editing based on those directives.
\textbf{2. Adapting to modifier directives during online motion generation.} 
In order to achieve the desired motion when modifier directives are given during task execution, 
it is necessary to map the modifier directives to robot actions and generate the next action. 
During online execution, the system generates smooth trajectories by applying a weighted average 
to the predicted action sequences.

In the proposed method, motions are divided into short action sequences, and weakly supervised labels 
reflecting the modifier directives are assigned to each sequence. During model training, the error 
between the latent variables and the weakly supervised labels, as well as the output, is backpropagated 
to constrain the learning of the latent representation, establishing the correspondence between 
the generated sequences and the modifier directives. Through this approach, the method simultaneously 
addresses the two challenges: linking modifier directives to latent representations and enabling online 
motion generation based on modifier directives.

The main contributions of this paper are as follows:
\begin{itemize} 
    \item We propose a motion generation model that can modify motions online based on modifier directives, 
    and demonstrate the applicability of latent representation learning for associating these directives with 
    motions, compared to existing imitation learning approaches. 
    \item By using a structure similar to Action Chunking for the predicted action sequences, we show that the desired motion can be achieved when modifier directives are input during online motion generation. 
\end{itemize}






%% file: src/02_rerated_work.tex
\section{Related Work}
\label{sec:citations}
\textbf{Imitation learning in robot manipulation.}
Imitation learning (IL) \citep{osaIL} is a fundamental approach in robot learning that trains robots from
human demonstrations. A representative example is Behavior Cloning (BC), and due to the simplicity of its 
learning process, a variety of architectures have been proposed. Traditionally, learning methods using
recurrent neural networks such as RNN \citep{RNN_ELMAN1990} 
and Long Short Term Memory (LSTM) \citep{LSTM_1997}, which can take 
temporal information into account, have been extensively studied. More recently, 
Transformer \citep{transformer_2017} based models, such as Action Chunking with 
Transformers (ACT) \citep{zhao2023}, have also been proposed. ACT generates a chunk of 
action sequences and achieves smooth motion by calculating a weighted average with past 
action histories. By extending ACT to incorporate language instructions as model inputs, 
methods enabling motion generation based on human instructions have also been 
proposed \citep{MTACT2023}. However, these methods mainly focus on selecting appropriate skills 
or tasks based on language instructions and do not explicitly handle motion generation 
that reflects adjustments specified by modifier directives.

\textbf{Motion Latent Space Manipulation.}
To efficiently learn the variations in human demonstration motions, it is effective to compress the 
demonstrations into low-dimensional information and learn their latent representations 
as distributions. Representative examples include studies embedding task skills into a 
latent space \citep{hausman2018, lynch2019}. Learning models often employ methods 
such as Variational AutoEncoder (VAE) \citep{vaepaper} 
and Conditional Variational AutoEncoder (CVAE) \citep{cvaepaper}, which assume distributions in the 
latent space. Several approaches have been proposed that constrain the learning of the 
latent space to acquire desired representations. These include representations 
related to physical constraints \citep{kutsuzawaRAL, newtonianVAEtaniguchi}, representations related 
to skills and motion primitives \citep{lynch2019, Noseworthy2019, Osa2020, skid_2021}, 
and representations related to motion styles \citep{hristov2021learning, pecan, SONG2023120625}. 
Among these, the studies \citep{hristov2021learning, pecan, SONG2023120625} are most closely 
related to this work, as they explore methods for learning interpretable or controllable 
latent spaces from demonstrations using weak supervision. However, these models primarily 
support feedforward motion generation, where motion is generated in a one-shot manner based 
on selected latent variables or styles, and they do not enable online modification or 
continuous editing of motion during execution.

%% file: src/03_proposed.tex
\section{Proposed Method}
\label{sec:Proposed Method}
The purpose of this research is to develop an online motion generation model that can respond 
to modifier directives. The proposed model takes as input the robot state $\bm{s}_t$ and a 
command value $\bm{z}$ representing the latent variable corresponding to the modifier directive, 
and generates motions dynamically according to these inputs. For example, in a wiping task, 
the robot state reflects information obtained through physical interaction with the board,
while the modifier directive specifies behavioral characteristics such as wiping 
strength (e.g., ``wipe strongly'' or ``wipe weakly'') or motion speed 
(e.g., ``wipe slowly'' or ``wipe fast'').

\begin{figure*}[t]
    \centering
    \includegraphics[width=12.5cm]{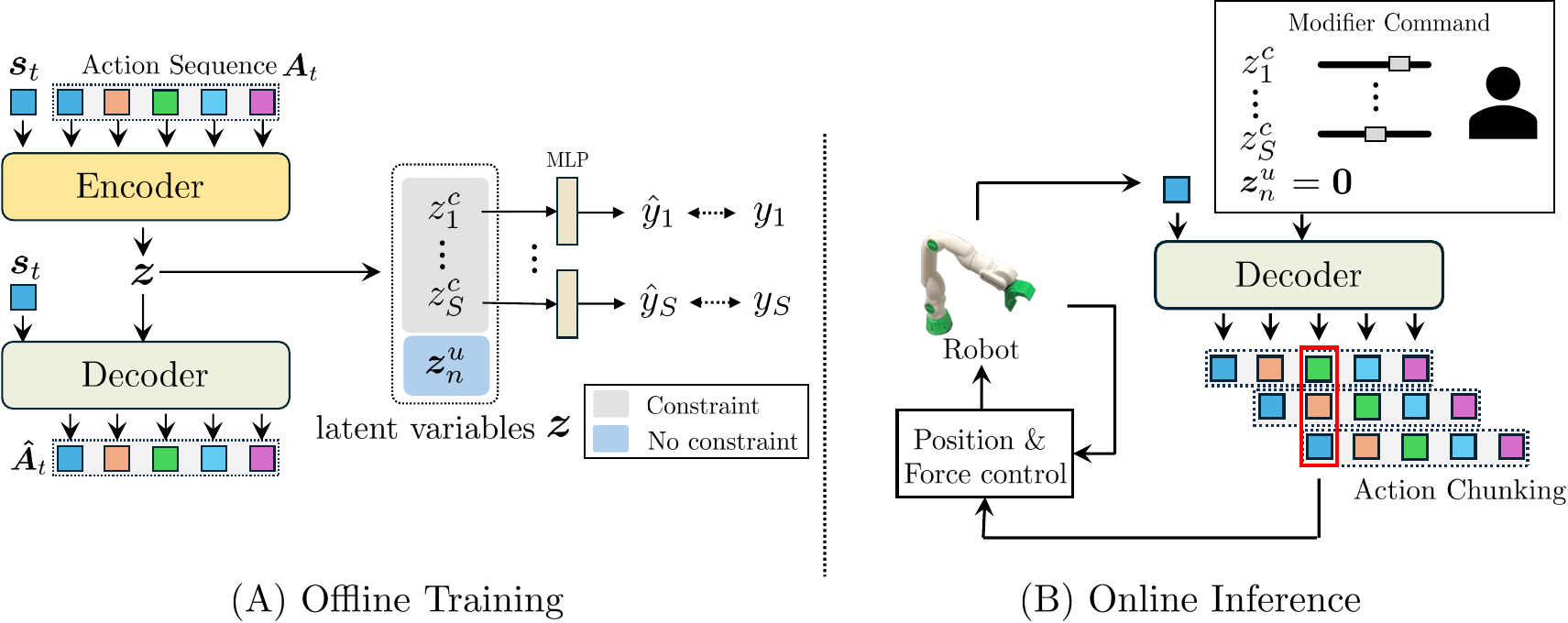}
    \caption{(A) Overview of the offline learning architecture. 
    (B) Overview of online inference.}
    \label{fig:overview}
\end{figure*}

Fig.~\ref{fig:overview} shows the overview of the proposed learning model. 
Fig.~\ref{fig:overview}(A) shows the offline learning architecture, while Fig.~\ref{fig:overview}(B) 
represents the overview of online inference. As shown in Fig.~\ref{fig:overview}(A), the proposed 
learning model primarily consists of two learning architectures: 
(1) an architecture that learns through CVAE
with action sequences $\bm{A}_t$ as input, and (2) an architecture for associating action sequences 
$\bm{A}_t$ with modifier directives. In the first architecture, a CVAE is used to build a motion 
generation model that learns the distribution of motion sequences starting from a specific initial 
state $\bm{s}_t$. In the second architecture, constraints are imposed on the CVAE's latent 
variables $\bm{z}$ to disentangle them according to the features of action sequences $\bm{A}_t$. 
This enables each disentangled latent variable to be trained to correspond to modifier directives. 
By editing variables in this learned latent space, motion generation that can control the robot
actions according to given modifier directives becomes possible.

\label{sec:proposed}
\subsection{Preliminaries}
\textbf{Data Collection Method}
To collect human motion data, we use a teleoperation method based on bilateral control.
Bilateral control involves two robots—a leader and a follower—where a human operates the 
leader robot, and the follower robot interacts with the environment. This setup enables the 
collection of motion data that accounts for the dynamics of the environment~\citep{saigusa}. 
In particular, four-channel bilateral control~\citep{sakainobilateral}, which simultaneously 
performs both position and force control on the leader and the follower, is effective for 
collecting data involving contact with the environment because it allows force information to 
be captured simultaneously~\citep{Sasagawa2020, kobayashiBiACT}. Its application to imitation learning 
has also been widely reported~\citep{kobayashiBiACT, yamane2024, tsuji2024mamba}.

\textbf{Robot System and Data Format}
For data collection and motion reproduction, we use the CRANE-X7 robot manufactured by RT 
Corporation. The CRANE-X7 has eight degrees of freedom, including the hand.
During motion recording, each joint's angle $\bm{q}_t$ [rad], angular velocity $\dot{\bm{q}}_t$ [rad/s], 
and torque $\bm{\tau}_t$ [N·m] at time step $t$ are recorded as the state 
$\bm{s}_t = [\bm{q}_t, \dot{\bm{q}}_t, \bm{\tau}_t]$.
A sequence of demonstrations is defined as a time series of robot states,
$\bm{\xi} = \{\bm{s}_1, \bm{s}_2, \dots, \bm{s}_T\}$,
where $T$ is the total number of time steps in the demonstration.
The collected motion data are organized into a dataset,
$\bm{\mathcal{D}} = \{\bm{\xi}_i\}_{i=1}^{M}$,
where $M$ is the number of demonstrations.

\subsection{Policy Learning}
In this section, we describe the learning architecture of the motion generation model.  
First, the collected motion data is divided into multiple action sequences  
$\bm{A}_t = \left[\bm{s}_t, \bm{s}_{t+1}, \dots, \bm{s}_{t+W-1}\right]$
using a sliding window of width $W$.  
The input to the learning model is the action sequence $\bm{A}_t$, and the output is 
the predicted action sequence $\hat{\bm{A}}_t$. In the CVAE, 
the encoder maps the action sequence $\bm{A}_t$  to the parameters of 
a probability distribution in the latent space.
The latent variable $\bm{z}$ is then sampled from this distribution, and the decoder 
learns to reconstruct $\bm{A}_t$ from this sampled $\bm{z}$.
The reconstruction error $\mathcal{L}_{rec}$ is defined as the mean squared error of the 
predicted state vectors and actual state vectors over the entire sequence of width $W$, 
as shown in equation (\ref{eqn:L_rec}), where $\hat{\bm{s}}_{t+j}$ denotes the predicted robot 
state at time $t+j$ generated by the decoder of the CVAE. The conditional variable for CVAE is 
the first sample of the action sequence $\bm{s}_t$, and the model learns the distribution 
of action sequences $\bm{A}_t$ starting from the initial state $\bm{s}_t$. 
The latent variable $\bm{z}$ is assumed to follow a diagonal Gaussian prior distribution 
$p(z)=\mathcal{N}(0, \mathbf{I})$.  
Thus, the KL regularization loss $\mathcal{L}_{kl}$, defined as the KL divergence from the 
approximate posterior $q_\phi (\bm{z} | \bm{A}_t, \bm{s}_t)$ 
to the prior $p(z)$, is shown in equation (\ref{eqn:L_kl}).

\begin{equation}
    \mathcal{L}_{rec} = \frac{1}{W} \sum_{j=0}^{W-1} (\bm{s}_{t+j} - \hat{\bm{s}}_{t+j})^2
    \label{eqn:L_rec}
\end{equation}
\begin{equation}
    \mathcal{L}_{kl} = D_{kl} \left[ q_\phi (\bm{z} | \bm{A}_t, \bm{s}_t) \ \| \ \mathcal{N}(0, \mathbf{I}) \right]
    \label{eqn:L_kl}
\end{equation}

\subsection{Disentangled Representation Learning Using Weakly Supervised Labels}
\begin{figure*}[t]
    \centering
    \includegraphics[width=12.5cm]{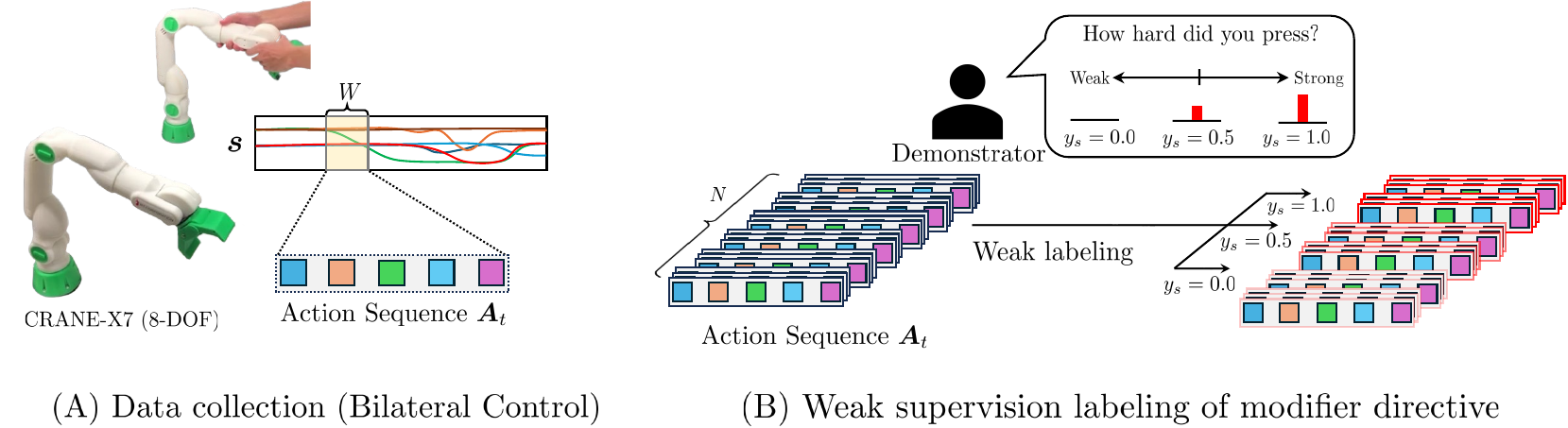}
    \caption{(A) Data collection through bilateral control. (B) Weakly supervised labeling of modifier directive by the demonstrator. In this example, labeling is performed for pressing force as follows: strong: $y_s=1.0$, moderate: $y_s=0.5$, weak: $y_s=0.0$.}   
    \label{fig:data collection}
\end{figure*}

In this section, we describe the method for associating the latent representations of CVAE 
with modifier directives. We assume $S$ types of modifier directives associated with the motion data 
and assign weakly supervised labels $y_s \in \{0.0, 0.5, 1.0\}$ for each type $(s = 1, 2, \dots, S)$. 
The values are ordered by directive intensity.
For example, in the case of a temporal directive representing motion speed, normal-speed motion is 
labeled with $y_s = 0.5$, fast motion with $y_s = 1.0$, and slow motion with $y_s = 0.0$. 
These modifier directives, such as ``fast'' and ``slow,'' are annotated by the demonstrator, 
as shown in Fig.~\ref{fig:data collection}.

To disentangle the behavioral characteristics associated with modifier directives embedded in human 
motion data, we impose constraints on a subset of the latent variables using weakly supervised labels.
We divide the latent variables into two groups: the constrained part $\bm{z}_s^{c}$ for 
modifier directives, and the unconstrained part $\bm{z}_n^u$, as defined in equation (\ref{eqn:z}). 
Here, $S$ denotes the number of modifier directives, and $N$ represents the number 
of unconstrained latent variables. Each constrained latent variable $\bm{z}_s^c$ is fed into a 
multi-layer perceptron (MLP) with ReLU activation, whose output is denoted by $\hat{y}_s$, 
and is trained to predict the corresponding weakly supervised label $y_s$. 
This process imposes supervision on $\bm{z}_s^c$ such that each dimension of the constrained 
latent variables encodes the demonstrator's intended modifier directives during training.
The training objective for this classification task is the binary cross-entropy loss 
$\mathcal{L}_{bce}$, as defined in equation (\ref{eqn:cel1}), where $\sigma(\cdot)$ denotes 
the sigmoid function.
For each modifier, the total loss is computed by summing the individual losses across all 
associated latent variables, as defined in equation (\ref{eqn:cel2}).
\begin{equation}
    \bm{z} =\{\bm{z}_{s}^c,\bm{z}_{n}^u\}
           =\{\underbrace{z_{1}^c, \dots, z_{S}^c}_{\text{constraint}},z_{1}^u, \dots, z_{N}^u\}
    \label{eqn:z}
\end{equation}
\begin{equation}
    \mathcal{L}_{s} =  \mathcal{L}_{bce}(y_s, \hat{y}_s) 
    = -[y_s\cdot\log(\sigma(\hat{y}_s))+(1-y_s)\cdot\log(1-\sigma(\hat{y}_s))]
    \label{eqn:cel1}
\end{equation}
\begin{equation}
    \mathcal{L}_{modi} =  \sum_{s=1}^{S} \mathcal{L}_{s}
    \label{eqn:cel2}
\end{equation}
\begin{equation}
    \mathcal{L} =  \alpha \mathcal{L}_{rec} +\beta \mathcal{L}_{kl} 
    + \gamma \mathcal{L}_{modi}
    \label{eqn:loss}
\end{equation}
The overall training loss $\mathcal{L}$ is computed as a weighted sum of the reconstruction 
loss $\mathcal{L}_{rec}$, the KL regularization loss $\mathcal{L}_{kl}$, and the modifier 
classification loss $\mathcal{L}_{modi}$, 
as shown in equation (\ref{eqn:loss}). The coefficients $\alpha$, $\beta$, and $\gamma$ are weighting 
parameters that control the contribution of each loss term. 
In other words, training is guided by a trade-off between learning the distribution of 
action sequences $\bm{A}_t$ from a given initial state $\bm{s}_t$ and predicting the appropriate 
modifier directives. Minimizing this objective encourages a latent representation that 
is disentangled with respect to the weakly supervised labels provided by the demonstrator.

\subsection{Online Inference}
As shown in Fig.~\ref{fig:overview}(B), during online motion generation, the robot state $\bm{s}_t$ 
and the modifier directives specified by a demonstrator are provided as inputs. 
The modifier directives are mapped to the latent representation 
$\bm{z} = \bm{z}_{\text{cmd}} = \{\bm{z}_s^c, \bm{z}_n^u\}$, which corresponds to the structure 
learned during training. The decoder then generates the action sequence $\hat{\bm{A}}_t$ 
with a length of $W(=50)$ based on these inputs. Among the latent variables, 
the unconstrained component $\bm{z}_n^u$ is fixed at $\boldsymbol{0}$, while the constrained 
component $\bm{z}_s^c$ is manually adjusted to reflect the desired modifier directives.
As shown in equation (\ref{eqn:WAC}), the next robot state $\hat{\bm{s}}_{t+1}$ is computed as a 
weighted average of the generated action sequence $\hat{\bm{A}}_t$ and previous generations.
Here $\hat{\bm{A}}_i[k]$ denotes the predicted robot state at relative step $k$ in the predicted 
sequence starting from time $i$. For $t=0$, our implementation applies a special-case handling 
without averaging and directly sets $\hat{\bm{s}}_{1}=\hat{\bm{A}}_{0}[1]$.
\begin{equation}
  \hat{\bm{s}}_{t+1}
  = \frac{\sum_{i=1}^{\min(t,\,W-1)} w_i\, \hat{\bm{A}}_{t+1-i}[i]}{\sum_{i=1}^{\min(t,\,W-1)} w_i},
  \qquad
  w_i = \frac{1}{\log(i+1)}.
  \label{eqn:WAC}
\end{equation}
The role of the weighted average with past outputs is not only to address the non-Markovian nature 
of motion, as highlighted in prior work \citep{zhao2023}, but also to mitigate abrupt transitions 
that may occur when $\bm{z}_{\text{cmd}}$ is modified during online motion generation. 
Furthermore, the robot is controlled via position and force control using the computed 
$\hat{\bm{s}}_{t+1}$ as the command signal. The details of the command signal update frequency 
and robot control are described in the APPENDIX \ref{sec:robot_config}. 
In this way, by providing command values $\bm{z}_{\text{cmd}}$ 
to the latent variables corresponding to modifier directives, the robot can autonomously 
generate motions that reflect the given directives.


%% file: src/04_experiment.tex
\section{Experiment}
\label{sec:experiment}
In the experiments, we provided $\bm{z}_{\text{cmd}}$ corresponding to the modifier directives 
as input and verified that the generated motions successfully reflected those modifier directives. 
Additionally, we demonstrate the effectiveness of Action Chunking in our method, where $\bm{z}_{\text{cmd}}$ 
is supplied during online motion generation. To validate these claims, we conducted two types of 
experiments. The first experiment assessed whether the generated motions were aligned with the given 
modifier directives in the task of wiping a whiteboard, utilizing two different types of modifier 
directives. The second experiment investigated the effectiveness of Action Chunking by comparing 
variations in the presence and type of weighted averaging applied in the proposed method.

\subsection{Evaluation Metrics}
We quantitatively assessed performance using two complementary metrics:
\textbf{1) Task Success Rate (TSR):} this metric evaluates whether the generated motion 
successfully completes the intended task. For each task, we defined specific success 
criteria to determine completion.
\textbf{2) Modifier Directive Errors (MDEs):} this metric quantifies the alignment 
between the modifier directives and the resulting motion. It measures the discrepancy 
between intended modifications and the generated motion across each latent variable dimension.
Detailed formulations for both metrics are provided in APPENDIX \ref{sec:eval_config}.

\subsection{Comparison Methods}
To evaluate our proposed approach, we implemented two distinct imitation learning architectures. 
The first is \textbf{ACT}~\citep{zhao2023}, a CVAE model using Transformer 
architecture for both encoder and decoder. The second comparison method is \textbf{CVAE-LSTM}, 
a CVAE model employing LSTM networks for both. Both methods incorporate 
Action Chunking on the generated sequence.
For our proposed extensions, we developed \textbf{ACT (Proposed)} and \textbf{CVAE-LSTM (Proposed)}, 
which extend the original models with a latent representation learning constraint. 
This constraint predicts weakly supervised labels from the latent 
variable $\bm{z}$ and backpropagates the error. By comparing each architecture with and without our 
proposed extension, we assess both the effectiveness of our approach and its adaptability to different 
models when responding to modifier directives during online execution.

\subsection{Experimental Setup}
\begin{figure*}[t]
    \centering
    \includegraphics[width=12.0cm]{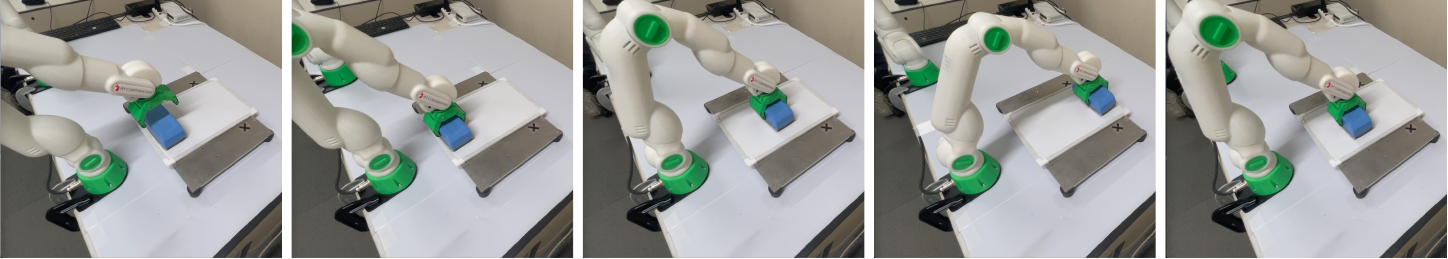}
    \caption{Wiping task: the robot grabbed the whiteboard eraser and uses its entire body 
    and joints to wipe the whiteboard.}    
    \label{fig:snapshot}
\end{figure*}

We evaluated our approach on the Wiping task, focusing on temporal and physical modifier directives. 
As illustrated in Fig.~\ref{fig:snapshot}, this task involves grasping a whiteboard eraser and 
wiping a whiteboard positioned at a fixed location. 
Temporal directives were defined based on the time required to complete three wiping cycles and 
evaluated under three conditions: Slow, Moderate, and Fast. Physical directives were characterized by 
the wiping pressure applied to the whiteboard, also evaluated under three conditions: Weak, Moderate, 
and Strong. Table~\ref{tab:modifier} shows the correspondence between these modifier directives 
and their associated weakly supervised labels.
The latent space dimension was set to 3, comprising $S=2$ latent variables constrained by weak supervision 
labels and $N=1$ unconstrained latent variable. In the Wiping task, $z_1^c$ corresponds to physical 
directives and $z_2^c$ to temporal directives, both constrained by labels during training.

\begin{table}[htbp]
    \centering
    \caption{Correspondence table between modifier directive and weakly supervised labels}
    \small
    \renewcommand{\arraystretch}{1.3}
    \begin{tabular}{c|c|c|c}
        \hline
         & $y_s=0.0$& $y_s=0.5$& $y_s=1.0$\\
        \hline
        \hline
         Physical (phys)& Weak & Moderate & Strong \\
        \hline
         Temporal (temp)& Slow & Moderate & Fast \\
        \hline
    \end{tabular}
  \label{tab:modifier}
\end{table}

\section{Results}
\label{sec:proposed}

\subsection{Motion Generation Using Modifier Directives}
To evaluate whether the generated motions accurately reflected the intended modifier directives, 
we computed the modifier directive errors (MDEs) for each latent variable based on the consistency between 
command variations and the intended directives. Table~\ref{tab:wiping_results} presents the Task Success 
Rate (TSR) and MDEs for the Wiping task. For the MDEs, lower values indicate that the latent 
variable more clearly represents the intended modifier. Latent variables with a compatibility score 
below 0.50 are highlighted in bold.

For the proposed method, we set the latent command to 
$\bm{z}_{\text{cmd}} = \{z_{1}, z_{2}, z_{3}\}=\{z_{1}^c, z_{2}^c, z_{1}^u\}$ 
and tested axis-aligned patterns in which exactly one constrained coordinate ($z_{1}^c$, $z_{2}^c$)
was set to $\pm \{ 1.0, 2.0\}$ while the others were fixed at 0.0; the unconstrained variable
$z_{1}^u$ was always 0.0 (9 settings, including a neutral setting of all zeros). 
This approach focuses the evaluation on the two dimensions 
explicitly constrained to represent modifier semantics.
In contrast, for the baseline methods (CVAE-LSTM and ACT), we used 
$\bm{z}_{\text{cmd}} =\{z_{1}^u, z_{2}^u, z_{3}^u\}$ and varied exactly one coordinate to
$\pm \{ 1.0, 2.0\}$ while fixing the other 
two at 0.0 (13 settings, including a neutral setting of all zeros). Since the baselines 
lack the proposed disentanglement constraint, it is not known which latent dimension 
corresponds to which modifier. Therefore, we evaluate all three dimensions.

\begin{table}[htbp]
  \centering
    \caption{Success rate and modifier directive conformity Index in the Wiping task}
  \small
  \begin{tabular}{l|c|c|c|c|c|c|c}
    \hline
    & Task success& \multicolumn{2}{c|}{$z_1$(phys)} & \multicolumn{2}{c|}{$z_2$(temp)} & \multicolumn{2}{c}{$z_3$} \\
    \cline{3-8}
    & rate (TSR)& temp & phys & temp & phys & temp & phys \\
    \hline\hline
    CVAE-LSTM & 100.0\% [13/13] & 1.09& 1.01& 1.06& 1.04& 1.11& 0.92 \\
    \hline
    \textbf{CVAE-LSTM (Proposed)} &  100.0\% [9/9] & 0.69& 0.63& \textbf{0.22}& 0.63& --& --\\
    \hline
    ACT &  0.0\% [0/13] & --& --& --& --& --& --\\
    \hline
    \textbf{ACT (Proposed)} &  88.9\% [8/9] &\textbf{0.43}& 1.47& 1.00&  1.95& --& --\\
    \hline
  \end{tabular}
  \label{tab:wiping_results}
\end{table}

The proposed method achieved higher task success rates compared to the baseline models, 
demonstrating its ability to generate motions aligned with the provided directives. 
Specifically, the standard CVAE-LSTM exhibited MDEs 
exceeding 1.00 across all latent variables, with minimal differentiation between 
$z_1$, $z_2$, and $z_3$. This suggests limited capability in associating individual 
latent dimensions with distinct directive types, indicating an entangled representation.
In contrast, CVAE-LSTM (Proposed) achieved substantially lower MDEs for the latent variables explicitly 
constrained by modifier directives, with the temporal MDE for $z_2$ dropping to 0.22 and the physical MDE 
for $z_1$ to 0.63. These reductions suggest that each latent variable more clearly 
and independently encodes a specific directive, supporting improved disentanglement 
through the proposed approach. Nevertheless, the temporal MDE for $z_1$ remained at 0.69, 
implying some residual entanglement between latent variables.
Similarly, ACT (Proposed) showed improved disentanglement compared to the original 
ACT, with a temporal MDE of 0.43 when varying $z_1$. However, the physical MDE for $z_2$ 
remained high (1.95). This is contrary to the intended behavior where $z_1$ should
represent temporal directives and $z_2$ should represent physical directives. This
suggests that modifier effects may still appear along unintended 
latent axes. This result highlights that, while the proposed modifications enhanced 
directive alignment, there remains room for improving the orthogonality and independence 
of latent variable representations.
These findings confirm that the proposed method can effectively generate motions that respond to 
temporal and physical modifiers in the Wiping task, demonstrating a clear correspondence between 
latent variables and modifier semantics.

\subsection{Effect of Action Chunking on Motion Stability}
To demonstrate the effectiveness of Action Chunking in our method, where $\bm{z}_{\text{cmd}}$ 
is supplied during online motion generation, we evaluated its impact on motion stability 
in the Wiping task. Table~\ref{tab:weight} summarizes the task success rate (TSR) under different 
temporal weighting functions in action chunking.
We compared the case without action chunking (No weight) against three weighting schemes.
When action chunking was not applied, the predicted actions exhibited severe oscillations 
during execution, resulting in a 0\% TSR.
In contrast, applying action chunking substantially improved motion stability and task 
performance. Notably, the use of $w_i = 1/\log(i+1)$ achieved a 100.0\% success rate for both 
the CVAE-LSTM and ACT variants of our method. This result highlights the importance of 
temporal smoothing in mitigating abrupt transitions in the predicted action sequence 
caused by changes in the latent command $\bm{z}_{\text{cmd}}$.
These findings empirically support the effectiveness of our proposed Action Chunking mechanism 
in enhancing online motion generation under modifier directives.

\begin{table}[htbp]
\centering
\caption{Relationship between weight parameters and task success rate (TSR) in action chunking}
    \renewcommand{\arraystretch}{1.2}
    \small
    \begin{tabular}{c|c|c|c|c}
        \hline
         & No weight & $w_i=1/(i+1)$ & $w_i=\exp{(-m*i)}$ & $w_i=\ 1 / \log(i+1)$ \\
        \hline
        \hline
        CVAE-LSTM (Proposed) & 0.0\% [0/5] & \textbf{100.0\%} [5/5] & 0.0\% [0/5] & \textbf{100.0\%} [5/5] \\
        ACT (Proposed) & 0.0\% [0/5] & 0.0\% [0/5] & \textbf{100.0\%} [5/5] & \textbf{100.0\%} [5/5] \\
        \hline
    \end{tabular}
    \label{tab:weight}
\end{table}

%% file: src/05_conclusion.tex
\section{Conclusion}
\label{sec:conclusion}
In this study, we present a motion generation model capable of dynamically adjusting 
robot behaviors in response to online modifier directives.
We propose a learning framework that links human-issued modifier directives with 
corresponding motion variations, enabling online adaptation of generated trajectories.
Our evaluation demonstrates that the model effectively handles specific types of directives, 
such as temporal and physical constraints.
Furthermore, by introducing an action chunking mechanism to mitigate abrupt transitions in motion 
sequences, we observe a notable improvement in task success rates.
These results validate the proposed method's effectiveness in generating responsive and 
adaptable motions under modifier directives.

%% file: src/06_appendix.tex
\newpage
\appendix
\section{LIMITATION}
\label{sec:limitation}
\begin{figure*}[htbp]
    \centering
    \includegraphics[width=11.0cm]{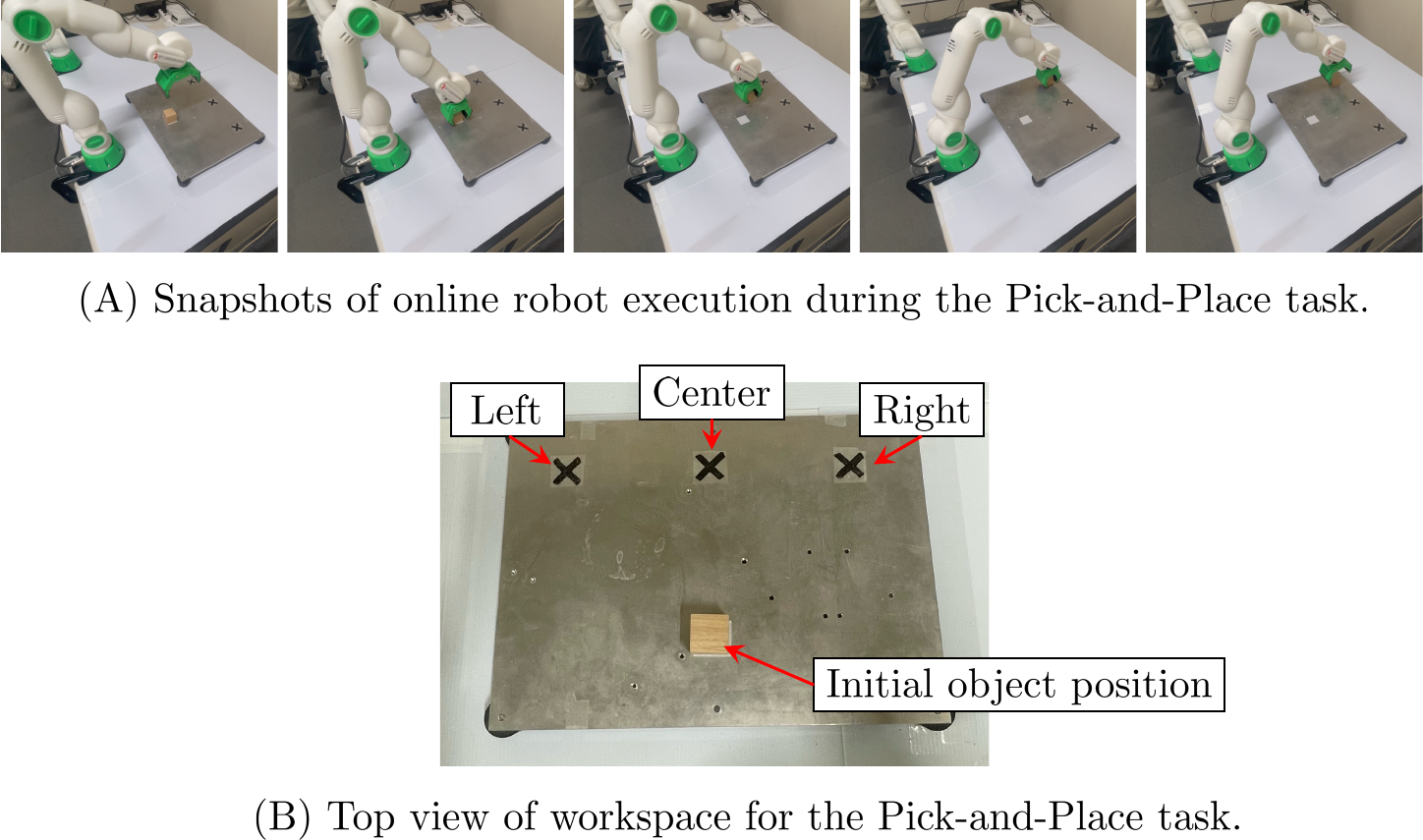}
    \caption{Pick-and-Place task: The robot picks up a block and places it at a designated location.}    
    \label{fig:pp_snapshot}
\end{figure*}

Although the proposed method enables motion generation that can respond to online modifier directives, 
there remain several limitations.
First, while our method demonstrates successful disentanglement between latent variables and 
modifier directives in certain tasks, this property does not generalize consistently across 
directive types or task domains. As an additional evaluation, we applied our framework to a 
Pick-and-Place task, where the robot is required to grasp an object and place it at a designated 
location, as shown in Fig.~\ref{fig:pp_snapshot}(A). This task involves two types of modifier 
directives: temporal (Slow, Moderate, Fast) and spatial (Left, Center, Right), with the latter 
specifying the object placement location, as shown in Fig.~\ref{fig:pp_snapshot}(B).
\begin{table}[htbp]
  \centering
  \caption{Success rate and modifier directive conformity index in the Pick-and-Place task}
  \small
  \begin{tabular}{l|c|c|c|c|c|c|c}
    \hline
    & Task success& \multicolumn{2}{c|}{$z_1$} & \multicolumn{2}{c|}{$z_2$} & \multicolumn{2}{c}{$z_3$} \\
    \cline{3-8}
    & rate (TSR) & temp & spat & temp & spat& temp & spat \\
    \hline\hline
    CVAE-LSTM  & 100.0\% [13/13] & 1.05& \textbf{0.05}& 1.02& 0.67& 1.01& 1.38 \\
    \hline
    \textbf{CVAE-LSTM (Proposed)} &  100.0\% [9/9] & 1.09& 1.04& 1.35& 2.06& --& --\\
    \hline
    ACT  &  100.0\% [13/13] & 1.00& 1.04& 1.05& 1.00& 0.98&1.04\\
    \hline
    \textbf{ACT (Proposed)} & 100.0\% [9/9] & 1.24& 0.97& 1.00&  1.00& --& --\\
    \hline
  \end{tabular}
  \label{tab:picking}
\end{table}

As shown in Table~\ref{tab:picking}, all models achieved high task success rates; however, 
the MDEs exceeded 1.00 for most directive-latent variable combinations.
A notable exception was observed in the CVAE-LSTM model, where varying $z_1$ yielded a very low spatial 
MDE of 0.05. Nevertheless, this effect was not consistently reproduced across other models or variables. 
These findings suggest that spatial directives—being symbolic and discrete in nature—are not well captured 
by the current framework, which is based on continuous latent variables and weak supervision.
This limitation highlights the need to develop more flexible mechanisms to accommodate the semantic 
nature of different directives. In particular, bridging the mismatch between discrete directive labels 
and continuous latent representations may require the introduction of additional architectural elements, 
such as gating mechanisms \citep{2019gatedvae} that explicitly handle symbolic task features.

Second, the weighting parameters $\alpha$, $\beta$, and $\gamma$ used in the loss function significantly 
affect training outcomes. In this study, we empirically tuned these weights individually for each task. 
However, the importance of each component—reconstruction, KL divergence, and modifier directive—may 
vary with the task type and directive semantics. Future work should explore task-adaptive or 
data-driven strategies to balance these components effectively.

Finally, our experiments were limited to proprioceptive input modalities: joint angles $\bm{q}$ [rad], 
angular velocities $\bm{\dot{q}}$ [rad/s], and joint torques $\bm{\tau}$ [N·m]. As a result, the 
evaluation was constrained to tasks where these modalities are the primary source of directive-related 
variation. In tasks such as polishing or inspection, directives like "carefully" or "thoroughly" 
may rely heavily on auditory or visual cues. Extending the framework to incorporate multimodal sensory 
inputs remains an important direction for future work.


\section{APPENDIX}
\label{sec:appendix}
\subsection{Robot Control Configuration}\label{sec:robot_config}
Bilateral control for data collection was executed with a control period of 2 ms, 
and the data were collected at this frequency. The data were then downsampled every 
20 steps and used for training.  
For autonomous operation, we followed the approach in  where commands are given to the 
follower robot. The follower robot performed position and force control with a control 
period of 2 ms, and the command values inferred by the model were updated every 20 steps.

\subsection{Evaluation Metrics Details}\label{sec:eval_config}
\paragraph{Task Success Rate (TSR).}
Task Success Rate (TSR) evaluates whether the robot successfully completes the assigned task, 
regardless of whether the generated motion reflects the intended modifier directive. 
For the Wiping task, a trial was considered successful if the robot could grasp 
the eraser from the initial pose and perform repeated back-and-forth wiping motions on the 
whiteboard. Failures were defined as the inability to grasp the eraser, premature stopping, 
or lifting the eraser off the whiteboard during execution. In the Pick-and-Place task, 
success was defined as grasping the object from the initial pose and placing it onto the 
designated workspace. Failure cases included missing the object during grasping, unintended 
gripper actuation mid-air, failure to open the gripper during placement, or unintentional stopping.

\paragraph{Modifier Directive Errors (MDEs).}
Modifier Directive Errors (MDEs) quantify the extent to which the generated motion aligns with a given 
modifier directive by comparing it to reference motions annotated with directive labels. For each 
directive type, we first collected reference motion data corresponding to three predefined levels 
(e.g., Slow, Moderate, Fast), as shown in Table~\ref{tab:modifier}. For each level, four trials 
were conducted, resulting in twelve reference samples. A task-relevant scalar feature was extracted 
from each trial and averaged across repetitions for each directive level. These three averaged 
feature values were then fitted to a line via least squares regression, yielding a reference line 
$y = ax + b$, where $x \in \{0.0, 0.5, 1.0\}$ represents the directive label normalized to the 
range $[0, 1]$, and $y$ is the corresponding feature value.

The fitted coefficients $a$ and $b$ used for each directive in both tasks are shown in 
Tables~\ref{tab:pick_and_place} and~\ref{tab:wiping}. For instance, in the 
Pick-and-Place task, the slope and intercept for the temporal directive were $-7.097$ 
and $16.525$, respectively, while for the spatial directive they were $-0.725$ and $3.526$. 
In the Wiping task, the temporal directive yielded a slope of $-6.379$ and intercept of $12.414$, 
and the physical directive produced values of $-1.115$ and $-1.017$.

In the experiment, the modifier directive was specified using five latent 
command values: $-2, -1, 0, 1, 2$. These values were linearly scaled to match 
the normalized directive space, resulting in $x \in \{0.0, 0.25, 0.5, 0.75, 1.0\}$. 
The corresponding feature values $y$ obtained from the generated motions were then 
fitted to a line $y = cx + d$ via least squares regression.

MDE is defined as the normalized Euclidean distance between the slopes and intercepts of the 
reference and generated lines, and is given by:
\begin{equation}
    \text{MDE}  = \sqrt{\left( \frac{a-c}{a} \right)^2 + \left( \frac{b-d}{b} \right)^2}
    \label{eqn:mde}
\end{equation}
To avoid division by zero, we assume $a \ne 0$ and $b \ne 0$.
Lower MDE values indicate better alignment between the intended directive and the generated motion. 
This metric is computed independently for each directive-latent variable pair to assess how well 
each latent dimension captures a specific type of directive.

The feature values $y$ used for each directive type are defined as follows. 
For the Wiping task, the average time required to complete three wiping cycles was used for temporal 
directives, while for the Pick-and-Place task, the time from the start of the motion to placing 
the object at the designated position was used. For Physical directives, we recorded the 
minimum torque $\tau_4 \, [\mathrm{Nm}]$ at joint 4 during each wiping cycle 
and computed the average over three cycles. For Spatial directives used in the Pick-and-Place task, 
we recorded the final joint angle $q_2 \, [\mathrm{rad}]$ of joint 2 at the completion of the placement action.

\begin{table}[htbp]
    \centering
    \caption{Reference line parameters $(a, b)$ for the Pick-and-Place task}
    \renewcommand{\arraystretch}{1.3}
    \begin{tabular}{c|c|c} 
        \hline
         Modifier Type & Slope ($a$) & Intercept ($b$) \\  
        \hline\hline
        Temporal & $-7.097$ & $16.525$\\  
        \hline
        Spatial & $-0.725$ & $3.526$\\  
        \hline
    \end{tabular}
    \label{tab:pick_and_place}
\end{table}

\begin{table}[htbp]
    \centering
    \caption{Reference line parameters $(a, b)$ for the Wiping task}
    \renewcommand{\arraystretch}{1.3}
    \begin{tabular}{c|c|c} 
        \hline
         Modifier Type & Slope ($a$) & Intercept ($b$) \\  
        \hline\hline
        Temporal & $-6.379$ & $12.414$\\  
        \hline
        Physical & $-1.115$ & $-1.017$\\  
        \hline
    \end{tabular}
    \label{tab:wiping}
\end{table}

\subsection{Learning Method Details}
Table~\ref{tab:experimental_settings} summarizes the training parameters and their values.
Table~\ref{tab:hyperparameters_act}, Table~\ref{tab:hyperparameters_cvae} and
Table~\ref{tab:hparams_stylelinear} show the 
hyperparameters used for the ACT model, the CVAE-LSTM model and the weak-label predictor, respectively.
The weights of each loss component are presented in Table~\ref{tab:loss_weights_all}.
The models were trained on a workstation equipped with an 11th Gen Intel (R) Core (TM) i7-11800H 
@ 2.30GHz CPU and an NVIDIA RTX A3000 Laptop GPU.
The overall training objective follows the loss formulation defined in Equation~\eqref{eqn:loss}, 
which combines the reconstruction loss $\mathcal{L}_{rec}$, the KL regularization loss $\mathcal{L}_{kl}$, 
and directive supervision loss $\mathcal{L}_{modi}$. The weights $\alpha$, $\beta$, and $\gamma$ 
control the contribution of each term and were determined empirically per task and model, as shown 
in Table~\ref{tab:loss_weights_all}. For baseline models (ACT and CVAE-LSTM), $\gamma$ was not applied 
since modifier supervision was not used.
\begin{table}[htbp]
    \centering
    \caption{Training parameters}
    \renewcommand{\arraystretch}{1.3}
    \begin{tabular}{ll}
        \hline
        \textbf{Parameter} & \textbf{Value} \\
        \hline
        Optimizer & Adam \citep{Kingma2014Adam} ($\beta_1 = 0.9$, $\beta_2 = 0.999$) \\
        Learning rate & $1.0 \times 10^{-3}$ \\
        Gradient clipping & 1.0 \\
        Batch size & 256 \\
        Epochs & 1000 (Seq2Seq), 5000 (ACT) \\
        \hline
    \end{tabular}
    \label{tab:experimental_settings}
\end{table}
\begin{table}[htbp]
    \centering
    \caption{Hyperparameters of ACT}
    \renewcommand{\arraystretch}{1.3}
    \begin{tabular}{ll} 
        \hline
        \textbf{Hyperparameter} & \textbf{Value} \\  
        \hline
        Encoder layers & 2 \\  
        Decoder layers & 3 \\  
        Hidden dimension & 48 \\  
        Feedforward dimension & 192 \\  
        Number of heads & 4 \\  
        Window size($=W$) & 50 \\  
        \hline
    \end{tabular}
    \label{tab:hyperparameters_act}
\end{table}
\begin{table}[htbp]
    \centering
    \caption{Hyperparameters of CVAE-LSTM}
    \renewcommand{\arraystretch}{1.3}
    \begin{tabular}{ll} 
        \hline
        \textbf{Hyperparameter} & \textbf{Value} \\  
        \hline
        Encoder layers & 3 \\  
        Decoder layers & 3 \\
        Output layer (fully connected)& 1 \\
        Hidden dimension & 256 \\  
        Window size($=W$) & 50 \\  
        \hline
    \end{tabular}
    \label{tab:hyperparameters_cvae}
\end{table}
\begin{table}[htbp]
    \centering
    \caption{Hyperparameters of the weak-label predictor}
    \renewcommand{\arraystretch}{1.3}
    \begin{tabular}{ll}
        \hline
        \textbf{Hyperparameter} & \textbf{Value} \\
        \hline
        Input dimension & 1 \\
        Hidden layers & 2 \\
        Hidden widths & [3, 3] \\
        Activation & ReLU \\
        Output dimension & 1 \\
        \hline
    \end{tabular}
    \label{tab:hparams_stylelinear}
\end{table}
\begin{table}[htbp]
    \centering
    \caption{Loss function weights ($\alpha$, $\beta$, $\gamma$) used in each task and model}
    \renewcommand{\arraystretch}{1.3}
    \small
    \begin{tabular}{l|l|c|c|c}
        \hline
        \textbf{Task} & \textbf{Model} & $\alpha$ & $\beta$ & $\gamma$ \\
        \hline \hline
        \multirow{4}{*}{Wiping} 
        & ACT & 1.0 & 4.0 & -- \\
        & ACT (Proposed) & 1.0 & 1.0 & 0.5 \\
        & CVAE-LSTM & 1.0 & 4.0 & -- \\
        & CVAE-LSTM (Proposed) & 1.0 & 0.3 & 2.5 \\
        \hline
        \multirow{4}{*}{Pick-and-Place} 
        & ACT & 1.0 & 4.0 & -- \\
        & ACT (Proposed) & 1.0 & 2.0 & 5.0 \\
        & CVAE-LSTM & 1.0 & 4.0 & -- \\
        & CVAE-LSTM (Proposed) & 1.0 & 0.3 & 2.5 \\
        \hline
    \end{tabular}
    \label{tab:loss_weights_all}
\end{table}